# Profiling of OCR'ed Historical Texts Revisited

Florian Fink
CIS
LMU
finkf@cis.uni-muenchen.de

Klaus U. Schulz
CIS
LMU
schulz@cis.uni-muenchen.de

Uwe Springmann
CIS, LMU
Humboldt-Universität zu Berlin
springmann@cis.uni-muenchen.de

## ABSTRACT

In the absence of ground truth it is not possible to automatically determine the exact spectrum and occurrences of OCR errors in an OCR'ed text. Yet, for interactive postcorrection of OCR'ed historical printings it is extremely useful to have a statistical profile available that provides an estimate of error classes with associated frequencies, and that points to conjectured errors and suspicious tokens. The method introduced in [3] computes such a profile, combining lexica, pattern sets and advanced matching techniques in a specialized Expectation Maximization (EM) procedure. Here we improve this method in three respects: First, the method in [3] is not adaptive: user feedback obtained by actual postcorrection steps cannot be used to compute refined profiles. We introduce a variant of the method that is open for adaptivity, taking correction steps of the user into account. This leads to higher precision with respect to recognition of erroneous OCR tokens. Second, during postcorrection often new historical patterns are found. We show that adding new historical patterns to the linguistic background resources leads to a second kind of improvement, enabling even higher precision by telling historical spellings apart from OCR errors. Third, the method in [3] does not make any active use of tokens that cannot be interpreted in the underlying channel model. We show that adding these *uninterpretable* tokens to the set of conjectured errors leads to a significant improvement of the recall for error detection, at the same time improving precision.

## Categories and Subject Descriptors

I.7.5 [**Document and Text Processing**]: Optical character recognition (OCR)—*Postcorrection*; I.2.7 [**Artificial Intelligence**]: Natural Language Processing—*German language*

## 1. INTRODUCTION

OCR engines of the latest generation employing recurrent neural networks with LSTM architecture lead to impressive results on OCR'ed historical documents over the complete history of modern printing [5, 4], often reaching recognition accuracy on the character level around 95% and more. Yet, for many applications OCR'ed texts need to be close to perfect, which means that postcorrection of OCR results remains inevitable. Fully automated postcorrection only helps to improve low quality texts. When already starting from a baseline accuracy of 95%, only interactive postcorrection leads to substantial improvements of OCR'ed texts, as automatic methods are prone to introduce new errors by way of miscorrections at a comparable or higher rate as real corrections.

The manual effort needed during interactive postcorrection can be substantially reduced if a *profiling tool* is available that points to suspicious tokens and conjectured errors and that indicates possible corrections in the text. In this way, a large number of errors can be found and corrected without having to scrutinize the complete document. At the same time, global estimates of the OCR error classes and frequencies in a document can be used for automated estimates of OCR quality, which is important when controlling output in large-scale digitization projects.

In [3] a profiling technology for historical OCR'ed documents has been introduced that offers these two functionalities. When detecting suspicious tokens and for calculating correction suggestions, historical spelling is taken into account using advanced language technology. The *global profile* computed for an OCR'ed historical text provides a ranked list of conjectured OCR error types with their expected frequencies, and similarly a list of historical patterns with expected frequencies in the text (*historical patterns* are rewrite rules such as $t \mapsto th$ that capture the typical differences in modern (t) and historical (th) spelling). For each OCR token, interpretations are generated that specify the (conjectured) OCR errors and patterns found in the token. In this way tokens with conjectured errors are immediately found (*local profile*). In [3] it has been shown that with the profiling method a good correlation between true and conjectured errors and error types is reached.

A weakness of the method in [3] is that the feedback during postcorrection provided by the manual corrections is not taken into account. The method is not adaptive and in the present form cannot use correction information to compute a refined profile with better statistical estimates. Here we show how the method can be modified to be open for this form of adaptivity. Manual corrections are used to simultaneously improve background lexica and estimates of the probability of OCR errors and historical patterns. Our experiments clearly confirm that the adaptive variant leads to superior profiling results for the unseen and uncorrected part of the document.

When inspecting the profiling results in several documents we found that another step helps to further improve profiling accuracy. While the spectrum of historical patterns used in [3] contains the differences between modern and historical German found in the IMPACT project[1], a closer look shows that additional patterns are found in documents from earlier periods not covered by IMPACT. After carefully inspecting our current collection of sources, an additional set of patterns has been collected. We show that with the enlarged set of patterns, better interpretations for OCR tokens are generated, which further improves profiling results.

When interpreting document profiling as a method for OCR error detection, the above methods (adaptivity and the extension of pattern sets) mainly improve *precision*. Our third improvement is related to the fact that several OCR tokens are *uninterpretable* under the profiling technology in the sense that the underlying channel model, which comes with fixed bounds on the number of OCR errors and patterns in a token, does not offer any possible interpretation for the token. In [3], uninterpretable tokens are completely ignored. We show that adding uninterpretable tokens to the set of possible errors leads to a *significant improvement of the recall* of error detection. Interestingly, a modest improvement of precision is achieved at the same time.

In our experiments we look at the postcorrection of two OCR'ed historical documents. In both cases, adaptivity, extension of pattern sets, and the use of uninterpretable tokens leads to a threefold improvement. For the interactive postcorrection, which provides the basis for adaptive improvements, we proceed in two ways. In a first series of experiments we correct all tokens of an initial part of the text. For evaluating the quality of profiles, this initial part part is ignored later on. In a second series of experiments we only correct tokens of the initial part that are marked as suspicious in the initial profile. The number of manual corrections in this series is much lower. Nevertheless, comparable improvements are obtained.

The paper is organized as follows. In Section 2 we sketch the profiling method introduced in [3]. Thereafter (Section 3) we introduce our modifications that lead to an adaptive profiling technology. Section 4 gives some information on the documents used for the evaluation. In Section 5 we describe the improvements that are achieved using the methods above. Section 6 comments on remaining problems. We finish with a short conclusion.

## 2. PROFILING OCR'ED HISTORICAL TEXTS

Consider a token $w_{\text{ocr}}$ delivered from an OCR engine for a historical text. Assuming that the OCR process did not confuse token borders, $w_{\text{ocr}}$ corresponds to a fixed word $w_{\text{gt}}$ of the underlying ground truth text. In the presence of OCR errors, $w_{\text{ocr}}$ and $w_{\text{gt}}$ may be distinct. In the absence of ground truth data we may have several candidate words $w_{\text{candgt}}$ that might represent the correct token of the text. The difference between $w_{\text{ocr}}$ and a candidate $w_{\text{candgt}}$ can be modeled in terms of an *OCR error trace* $\tau_{\text{ocr}}$, i.e., a sequence of error operations applied to $w_{\text{candgt}}$ and leading to $w_{\text{ocr}}$. For each of the different candidate words $w_{\text{candgt}}$ we have a pair $(w_{\text{candgt}}, w_{\text{ocr}})$ and a (hypothetical) OCR error trace.

If we want to use a lexicon of modern language to validate a token $w_{\text{gt}}$ or $w_{\text{candgt}}$ we need to take into account that modern and historical spelling might be distinct. Assume that the typical differences between modern and historical spellings have been captured in terms of a set of historical patterns *Pat* (local rewrite rules). Then the correspondence between a historical word $w_{\text{candgt}}$ and a possible modern equivalent $w_{\text{candmod}}$ can be described as a *historical trace* $\tau_{\text{hist}}$, a derivation where rules from the set *Pat* are applied to rewrite a word $w_{\text{candmod}}$ from the modern background lexicon into $w_{\text{candgt}}$. Note that the usage of a modern lexicon of wordforms together with a set of historical rewrite rules is just a substitute for a full historical lexicon underlying the printed document at hand which we generally do not possess, as it would have to reflect the printing period, local dialects, and document intrinsic spelling variations.

[1] http://www.digitisation.eu/

Combining both aspects we obtain a *two-channel-model* where a modern word $w_{\text{mod}}$ is first rewritten into a historical equivalent $w_{\text{gt}}$, and afterwards $w_{\text{gt}}$ is rewritten into the OCR result $w_{\text{ocr}}$ described by a historical trace $\tau_{\text{hist}}$ and an OCR trace $\tau_{\text{ocr}}$: $w_{\text{mod}} \xrightarrow{\tau_{\text{hist}}} w_{\text{gt}} \xrightarrow{\tau_{\text{ocr}}} w_{\text{ocr}}$. Following [3] we define a *full interpretation* for an OCR token $w_{\text{ocr}}$ as a quintuple written as

$$w_{\text{candmod}} \xrightarrow{\tau_{\text{hist}}} w_{\text{candgt}} \xrightarrow{\tau_{\text{ocr}}} w_{\text{ocr}}$$

where $w_{\text{candgt}}$ is a possible ground truth version of $w_{\text{ocr}}$, $w_{\text{candmod}}$ is a possible modern equivalent for $w_{\text{candgt}}$, and $\tau_{\text{ocr}}$ and $\tau_{\text{ocr}}$ are traces of the above form. A *ground truth interpretation* for $w_{\text{ocr}}$ is a triple

$$w_{\text{candgt}} \xrightarrow{\tau_{\text{ocr}}} w_{\text{ocr}}.$$

As an example, a possible interpretation for the OCR token *tnurm* found in an OCR'ed historical German text is

$$\textit{Turm} \xrightarrow{T \mapsto th} \textit{thurm} \xrightarrow{h \mapsto n} \textit{tnurm}.$$

The profiler only looks at OCR tokens with more than three letters. In practice, a bound is put on the maximal number of pattern applications and OCR errors to avoid a huge number of irrelevant interpretations. This means that tokens may be *uninterpretable*.

For each interpretable token of the OCR output, the *profiling method* described in [3] generates a set of weighted interpretations. The basis is a matching procedure [2] that efficiently generates all possible interpretations for an OCR token, given a pattern set, a modern lexicon, and a bound for the maximal number of OCR errors and patterns in traces.

In the following we first sketch how weighted interpretations are generated for each OCR token and afterwards describe the full profiling procedure. For details we refer to [3].

The background resources used for a specific historical language (e.g., historical German, or historical French) are

1. a lexicon $L_{\text{mod}}$ of modern words (full forms),

2. a set of historical patterns *Pat* for the specific language (German, French),

3. a lexicon $L_{\text{hist,traced}}$ of words in historical spelling with historical traces and modern equivalents checked and validated by a linguist,

4. a lexicon $L_{\text{hist,untraced}}$ of words in historical spelling without validated historical traces and modern equivalents,

5. an additional lexicon $L_{\text{spec}}$ with special vocabulary (names, geographical locations, Latin expressions, etc.).

All lexica are disjoint. While resources 1, 2 are necessary, resources 3-5 help to improve results. Words in $L_{\text{hist,traced}}$ and $L_{\text{hist,untraced}}$ come from a proofread historical corpus.

A *cascaded approach* is used to compute interpretations for a token $w_{\text{ocr}}$. First, if $w_{\text{ocr}}$ is in $L_{\text{mod}}$, only a single trivial interpretation $w_{\text{candmod}} = w_{\text{candgt}} = w_{\text{ocr}}$ with two empty traces is introduced.

Second, if $w_{\text{ocr}}$ is in $L_{\text{hist,traced}}$, it has been assigned a modern equivalent and a historical trace. We let $w_{\text{candgt}} = w_{\text{ocr}}$ and add an empty OCR trace. If $w_{\text{ocr}}$ is in $L_{\text{hist,untraced}}$, then we also define $w_{\text{candgt}} = w_{\text{ocr}}$ and the aforementioned matching procedure is used to generate modern equivalents and historical traces.

Third, if exact lexicon lookup against modern and historical lexica does not lead to any match, then the full matching procedure is used to search for interpretations with non-empty OCR traces and historical patterns using the pattern set *Pat*. Also, at this stage the additional special lexicon $L_{\text{spec}}$ gets used in the full two-channel matching procedure. Search is restricted in the sense that a small number (2-3) of pattern applications (or two OCR errors) are tolerated in the historical (or OCR) trace.

As a base score, we assign to each interpretation the product of the probabilities of the operations in the two traces with $P(w_{\text{candmod}})$. A smoothed unigram model for modern language is used to estimate $P(w_{\text{candmod}})$. We then distribute the *probability mass* 1 among all interpretations of an OCR token, using base scores of interpretations as proportionality factors.

A *text-channel-model* [3] for an OCR'ed input text $T$ is a triple $(V, O, H)$ where $V$ is a probability distribution for the set of words in $L_{\text{mod}}$ that estimates the probability for a modern word $w_{\text{mod}}$ to occur in the *true* trace of an OCR token from $T$. $O$ is a probability distribution which defines the probability for each OCR error in a set of possible errors. $H$ is a probability distribution which defines the probability for each pattern in *Pat*.

The profiling procedure for an OCR'ed input test starts with a naive text-channel-model $(V_0, O_0, H_0)$ for the OCR'ed input test $T$. Uniform probabilities for historical patterns and for OCR error types are used at this initialization step.

The procedure is organized in rounds. At round $n$, using the actual text-channel-model $(V_n, O_n, H_n)$ and the above cascaded method for generating interpretations we obtain a spectrum of weighted interpretations for all interpretable OCR tokens in $T$. Accumulating the probabilistic information contained in all these interpretations in an appropriate way, word, pattern, and OCR error probabilities are re-estimated to obtain a refined text-channel-model $(V_{n+1}, O_{n+1}, H_{n+1})$. The procedure stops after a number of rounds (we use 4 rounds, as we have found that more rounds do not improve the results) when differences between models become negligible. The complete procedure can be considered as a special expectation maximization (EM) algorithm.

Note that for each interpretable token $w_{\text{ocr}}$ we then have a set of weighted interpretations. In this way we have a conjecture which tokens are erroneous and how they can be corrected.

## 3. ADAPTIVE PROFILING

In order to make the above procedure adaptive, we need to modify the generation of weighted interpretations for the tokens $w_{\text{ocr}}$ that have been *corrected* by the user. The word *correction* is slightly misleading: in some cases, the user may just have added the information that in fact $w_{\text{gt}} = w_{\text{ocr}}$. In other situations the word $w_{\text{gt}}$ specified by the user is distinct from $w_{\text{ocr}}$.

In terms of the two-channel-model mentioned above and the generation of weighted interpretations, the difference for *corrected* tokens is that there is no need to speculate about the correct word among all candidates $w_{\text{candgt}}$ - we already have the correct word $w_{\text{gt}}$! As a first consequence, when looking for an optimal OCR trace of $w_{\text{gt}}$ it does not make sense to delimit the number of OCR errors. For the OCR part, we just compute an optimal OCR trace $\tau_{\text{ocr}}$ for $w_{\text{gt}}$ and $w_{\text{ocr}}$, allowing for an arbitrary number of errors and using the probabilities of the OCR errors in the current text-channel-model $(V_n, O_n, H_n)$.

Let us now look at the historical channel. Here again we use a cascaded approach.

1. If the user-corrected token $w_{\text{gt}}$ is in $L_{\text{mod}}$, then we set $w_{\text{candmod}} := w_{\text{gt}}$ and use a single interpretation $w_{\text{gt}} \xrightarrow{[]} w_{\text{gt}} \xrightarrow{\tau_{\text{ocr}}} w_{\text{ocr}}$ with an empty historical trace indicated by empty brackets.

2. If $w_{\text{gt}}$ is in $L_{\text{hist,traced}}$, it has been assigned a modern equivalent $w_{\text{mod}}$ and a historical trace $\tau_{\text{hist}}$. We use a single interpretation $w_{\text{mod}} \xrightarrow{\tau_{\text{hist}}} w_{\text{gt}} \xrightarrow{\tau_{\text{ocr}}} w_{\text{ocr}}$. If $w_{\text{gt}}$ is in $L_{\text{hist,untraced}}$, then the aforementioned matching procedure is used to generate modern equivalents and historical traces. For matching, we do not need to consider OCR errors because we have $w_{\text{gt}}$, and the maximum number of pattern applications is three.

3. If $w_{\text{gt}}$ is not in any of our dictionaries, the matching procedure is used to search for possible modern equivalents and historical traces. At this point there are two possible outcomes. (a) If we find possible modern equivalents $w_{\text{candmod}}$ and historical traces $\tau_{\text{hist}}$, then each such pair leads to an interpretation $w_{\text{modcand}} \xrightarrow{\tau_{\text{hist}}} w_{\text{gt}} \xrightarrow{\tau_{\text{ocr}}} w_{\text{ocr}}$. (b) If, however, we do not find any possible modern equivalent, then we introduce a *token without historical interpretation* and a partial trace $w_{\text{gt}} \xrightarrow{\tau_{\text{ocr}}} w_{\text{ocr}}$. In both cases (a) and (b), after having generated these (partial) interpretations, the token $w_{\text{gt}}$ is added to the lexicon $L_{\text{hist,untraced}}$ of known historical words.

The addition of $w_{\text{gt}}$ to $L_{\text{hist,untraced}}$ is important since it affects the possible interpretations of tokens from this point on.

The assignment of probabilities to the different interpretations for a corrected token is as before (if the token does not have a historical interpretation, only OCR pattern probabilities are taken into account). In the $n$-th round of the profiling the recalculation of probabilities for words, historical patterns and OCR errors remains as before, with a simple modification: for recalculating *word* and *pattern* probabilities, all tokens without a historical interpretation are ignored.

## 4. EVALUATION DATA AND PRINCIPLES

We tested on two German texts from the 16th and 17th century printed in Gothic typefaces:

- Adam von Bodenstein, *Wie sich meniglich...*, Basel 1557[2] (XXVIII recto to XLVII verso), and

- Bartholomäus Carrichter, *Kräutterbuch...*, Straßburg 1609[3] (pp. 47-75),

denoted in the following by 1557-W and 1609-K, respectively. Ground truth for these texts is available from the RIDGES corpus [1].[4] Each text contains about 5,000 transcribed tokens.

Text recognition was achieved using OCRopus with a model trained on a corpus of ten other German books printed from the 16th to 19th century (for details, see [5]). Character (*word*) recognition accuracy is 93.14% (*72.99%*) for Bodenstein and 97.34% (*90.74%*) for Carrichter.

To show the effect of a historical pattern list adapted to a document at hand, we employ two different lists: A basic list and an extended list containing 145 and 201 patterns, respectively. The basic list, which goes back to the IMPACT project, contains the most frequent patterns such as *s:f*, *u:v*, consonant doublings such as *n:nn* etc. The extended list was built by looking at previous profiler output in the context of our postcorrection tool PoCoTo[5] [6], when apparent prominent OCR error patterns turned out to actually represent an additional historical pattern. In this way we found historical spelling patterns such as *ß:ʃʃ* (see Fig. 1) which the profiler had treated as an OCR error pattern and therefore proposed to correct it to its modern spelling with *ß*. Additional patterns contain characters with diacritical marks oder superimposed vowels (e.g., ñ, ó, ů).

---

[2] http://reader.digitale-sammlungen.de/de/fs1/object/display/bsb11106588_00064.html
[3] http://reader.digitale-sammlungen.de/de/fs1/object/display/bsb10727266_00071.html
[4] Lüdeling, Anke; Odebrecht, Carolin; Zeldes, Amir; RIDGES-Herbology (Version 5.0), Humboldt-Universität zu Berlin, https://www.linguistik.hu-berlin.de/en/institut-en/professuren-en/korpuslinguistik/research/ridges-projekt?set_language=en
[5] https://github.com/cisocrgroup/PoCoTo

Figure 1: PoCoTo concordance view of OCR error series

Table 1: Baselines for 1557-W

| baseline | precision | recall | tp | fp | tn | fn (fair) | fn (obj) |
|---|---|---|---|---|---|---|---|
| 0 | 0.456 | 0.825 | 696 | 829 | 2797 | 148 | 646 |
| 1 | 0.461 | 0.841 | 710 | 831 | 2795 | 134 | 632 |
| 2 | 0.468 | 0.869 | 733 | 835 | 2791 | 111 | 609 |
| 3 | 0.469 | 0.871 | 735 | 834 | 2792 | 109 | 607 |

Note that the number and type of patterns depend on the level of *diplomatic accuracy*, that is the extent to which a gold truth transcription respects the printed glyphs. E.g. our historical lexica from the IMPACT project do not contain any long s, whereas our gold truth do, so the *ʃ:s* pattern is a real historical pattern in our data but tokens with a long *s* ( ʃ ) cannot be matched directly against the historical lexica. Nevertheless, these patterns will still be recognized by our profiler as historical patterns and not be treated as errors (Sec. 5).

In our experiments we split each OCR text into two parts. An initial segment consisting of 20% of all OCR tokens was used to imitate user corrections in order to study the effect of adaptivity. 80% of the texts were left untouched during these correction steps and used later to evaluate the quality of adaptive profiles obtained. All profiles were only evaluated on the unseen 80% part of the texts.

The quality of the profiles as an error detection mechanism was measured in the following way: Two ways of how to define the set of *predicted error tokens* were used. In the first setting, predicted error tokens are defined as the interpretable tokens $w_{ocr}$ in the 80% evaluation part where the top-ranked profiler interpretation for $w_{ocr}$ has a non-empty OCR trace (alternative interpretations with lower ranks are ignored). In the second setting, in addition all uninterpretable alphabetical tokens $w_{ocr}$ of length $> 3$ in the 80% evaluation part are treated as predicted error tokens. The available ground truth $w_{gt}$ for $w_{ocr}$ is used to define *true OCR error tokens* ($w_{gt} \neq w_{ocr}$).

*Precision* of error detection is the percentage of true OCR error tokens among all predicted error tokens. *Objective recall* is the percentage of predicted error tokens among all true error tokens. Since the profiler only looks at alphabetic tokens of length $> 3$, we mainly consider *fair recall*, which is defined as the percentage of correctly predicted error tokens among all true error tokens inspected by the profiler.

**Token migration analysis.** To get a better understanding of the effects of our various experiments, we do not just want to look at overall ratios such as precision and recall, but we also want to inspect what happens to individual OCR tokens, how many are reclassified and in which way. We therefore introduce the method of *token migration analysis* based on the $2 \times 2$ contingency table known from information retrieval, here with dimensions *profiler classification* (*positive*: erroneous OCR token; *negative*: correct OCR token) and *state-of-the-world* (profiler classification is *true* or *false*). Each OCR token gets therefore classified as either true-positive (*tp*), false-positive (*fp*), true-negative (*tn*), or false-negative (*fn*), with the sum of all four classifications being constant and equal to the number of OCR tokens. In view of the definitions given earlier, *precision* (percentage of retrieved tokens that are erroneous) is then *tp/(tp+fp)* and *recall* (percentage of erroneous tokens retrieved) is *tp/(tp+fn)*. The difference between any two profiling methods therefore consists of a redistribution of tokens among these four compartments with the restriction that reclassifications can only happen diagonally: Because the state-of-the-world is unchanged, a changed profiler classification (*positive-negative*) will also change the external assessment (*true-false*), so transitions only happen between *fp ↔ tn* and *fn ↔ tp*.

Since token number is conserved under classifications, the sums *fp+tn* and *fn+tp* are also separately constant. Table 1 gives an example of a token classification illustrating the above statements.

## 5. EVALUATION RESULTS

The optimization techniques discussed above can be grouped into three categories. *Adaptivity* means to take user feedback into account for computing refined profiles after correcting some tokens. When enlarging the set of patterns, the *background resources* are addressed. When adding uninterpretable tokens to the set of predicted errors, we do not modify the profiler but the *error predicting decision mechanism*. The latter two aspects are covered in the first evaluation part.

**1. Four baselines.** Following the distinct profiling strategies mentioned in the introduction and first ignoring adaptivity, not using any kind of user feedback, four baselines for precision and recall are obtained. As our main point of departure, *Baseline 0* (original method, *blue column* in Figs. 2 and 3) represents the values obtained when using the original profiling method from [3]. For *Baseline 1* (using enlarged pattern set, *red*) the enlarged set of patterns has been used for generating interpretations. *Baseline 2* (adding uninterpretable tokens, *yellow*) is obtained using the original set of patterns, adding uninterpretable tokens to the set of predicted error tokens. Finally *Baseline 3* (uninterpretable tokens & enlarged pattern set, *green*) is obtained when using the enlarged set of patterns, at the same time adding uninterpretable tokens to the set of predicted error tokens.

The resulting values of precision and recall for these four profiling methods are shown in Figs. 2 and 3. The color-coded recall values correspond to *fair recall (including tokens with length > 3*, and the corresponding values for *objective recall* (including all tokens) are shown as grey columns. Additionally, Table 1 shows the number of tokens for each token classification for 1557-W.

*Both precision and recall increase over the four profiling methods with a specifically large increase due to the inclusion of uninterpretable tokens into the set of error candidates (method 2).* This shows the general applicability of our procedure. The following section gives an in-depth interpretation of the single processes that cause this behavior.

**Detailed interpretation.** The best way to understand the effects of the various profiler methods on precision and recall is to look at the token migrations they give rise to. From Fig. 2 and Table 1 we see that the increase in recall appearing from method 0 to 1 is mostly due to an increased number of true positives. This is the result of uninterpretable tokens (false negatives) that become available to an error interpretation (Levenshtein distance < 3) once a new historical pattern has brought these tokens into reach. An example is the token ſpteiſſel that does not have an error interpretation with regard to the lexical token, not even to a token of a hypothetical lexicon generated by a modern lexicon with application of historical patterns. The modern equivalent is *spreißel (splinter)*, and the historical form *ſpreiſſel* is the ground truth. As we do not allow to apply historical patterns on top of other historical patterns, we would end up (employing *ß:ss* and *s:ſ* in first position) with *spteissel* and would now need three OCR errors to match the lexical word. However, with the new pattern list containing the pattern *ß:ſſ*, an interpretation with two historical patterns and one OCR pattern becomes available, so this token migrates from *fn* to *tp*.

**Figure 2: 1557-W: Baselines**

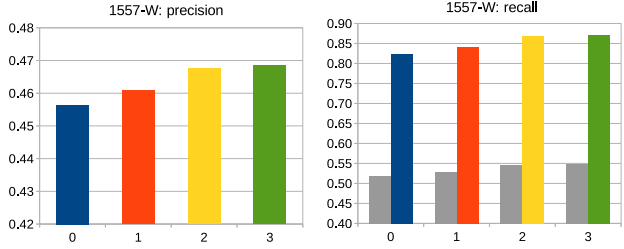

**Figure 3: 1609-K: Baselines**

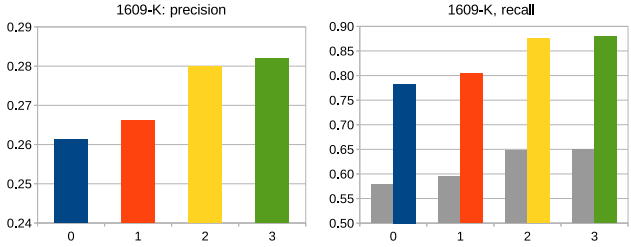

**Figure 4: 1557-W: Adaptive**

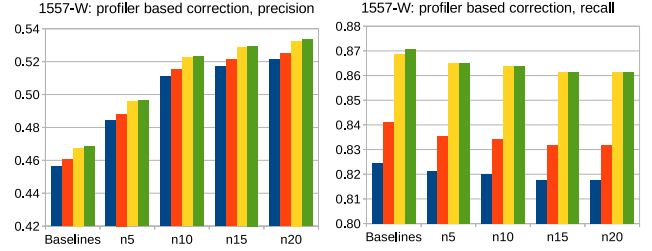

**Figure 5: 1609-K: Adaptive**

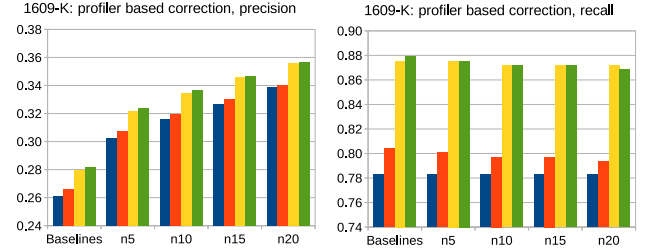

Note that uninterpretable tokens could also migrate from *true negatives* to *false positives*. In this case, recall would not be affected, but precision would suffer. As we can see in Table 1, this effect is of minor importance, because precision increases as well. An example is the correctly recognized OCR token *weißfărbig (white colored)*, which becomes interpretable after the pattern *ä:å* is introduced. Because the closest token in the lexicon is *weißfärbung*, the interpretation also contains an OCR error and therefore causes a migration from *true negative* to *false positive*.

A secondary effect of extended pattern lists consists in making historical interpretations possible which compete with previous error interpretations. This is a difficult case where the historical pattern channel and OCR error channel are both possible and cannot be unambiguously told apart. Which interpretation will dominate depends on the pattern list and initial pattern and OCR probabilities. Because for our classification purposes we only look at the most probably interpretation (treating the profiler as an error *detection* tool), the resulting classification is therefore somewhat arbitrary; for the purpose of generating *correction* candidates, however, the correct interpretation may still show up further down in the list of candidates. At any rate, this effect will convert positive tokens to negative ones ($tp \to fn$ or $fp \to tn$).

The observable change in positive and negative tokens is then the result of both the primary positive-enriching and the secondary positive-depleting effects. A single change in *tp* will change recall, whereas precision also depends on changes in *fp*.

Next comes the effect of treating all unidentifiable tokens as positive ones (method 1 to 2). This mostly dominates the positive-enriching effect of additional patterns, because now even those uninterpretable tokens become positive which under historical patterns would have stayed negative. As before, there is an increase in *true positives* and a smaller increase in *false positives*. The only additional increase in *true positives* from method 2 to 3 happens in the case where we had a token with a purely historical interpretation that later switches to another interpretation involving an OCR error. This is a rare event in our data, however, and vanishes completely with our adaptive methods which seem quite robust in solving disambiguities between both channels (historical and OCR) competing for the topmost interpretation. Some decrease in *false positves* still happens from method 2 to 3 in the event when a previous uninterpretable token happend to be correct (*true negative*), get included in positives (*false positive*) and later becomes interpretable because of a new historical pattern (again, *true negative*). This is the case for *Schólkraut, Súßholtz* in 1609-K. Because *tp* and consequently *fn* stay essentially constant, so does recall; precision still rises because of falling *fp*.

*The main route of token migrations from methods 0 to 3 is therefore from* fn *to* tp*, making previously unterpretable error tokens available to error interpretations and consequently increasing both precision and recall.*

Comparing both books, both precision and recall are lower for 1609-K than for 1557-W. This is due to the fact that 1609-K has a much better word recognition accuracy (91%) than 1557-W (73%). It is therefore much harder to detect the remaining errors in 1609-K which represent the long tail of an error distribution consisting mainly of single unrelated errors.

**2. Adaptivity (full correction).** In the main part of the evaluation we study the effect of adaptivity. As for the baselines, four methods for computing profiles and error prediction were compared (original method, using enlarged pattern set, adding uninterpretable tokens, adding uninterpretable tokens & enlarged pattern set). We considered four stages of user feedback, respectively correcting 25%, 50%, 75%, and 100% of the initial parts of the texts (i.e., 5%, 10%, 15%, and 20% of the full texts, denoted by *n5, n10, n15, n20* in the following). At each stage, a profile for the full text was computed using the adaptive method described in Section 3. Evaluation refers to the quality of profiles on the 80% parts of the documents.

Because the results for full correction were essentially the same as for profiler-based corrections (next section), the description and interpretation of the results will be given there.

**3. Adaptivity (profiler-based correction).** In a parallel series of experiments we did not use all tokens of the 5%, 10%, 15%, and 20% of the full texts for generating user feedback, but only those tokens of the respective parts that were marked as suspicious by the profiler. The results for precision and recall are given in Fig. 4 and 5. The fact that full correction and profiler-based correction almost lead to the same adaptive improvements for profiling of other text parts yet in another way sheds light on the value of the profiler technology.

The same reasoning as for the baselines applies to our experiments in adaptivity; again the color coded columns correspond to the different methods employed). For each column group denoted by the amount of additional input from a human corrector, precision and recall rise over methods 0 to 3. The more information is added, the higher the values become. *The effect of adding external information by corrections leads to increasing levels of precision which additionally rises for each group from methods 0 to 4. Recall stays essentially constant with a very slight decrease of about 1 percentage point.* Adaptivity therefore helps to save correction time, but does not increase the number of detectable errors.

**Detailed interpretation.** Looking at the effect of adding more information for each method separately, there is an almost constant number of *tp* (only decreasing by a few tokens) and a strongly decreasing number of *fp*, leading to a slight decrease of recall and a constant increase in precision. The learnings from corrections enlarge the historical lexicon and give preference to historical interpretations competing with OCR error interpretations. In this way, adaptivity can overcome the previously noted ambiguity in cases having both an interpretation as OCR errors and historical patterns.

*The main effect of adaptivity is the detection of* false positive *tokens with a consequent rise in precision and almost unaffected recall.*

## 6. ANALYSIS OF UNSOLVED PROBLEMS

From a postprocessing standpoint the more important measure is recall: If our profiler would be able to lead the user to most of those relatively rare tokens that are in error (even if some turn out to be false positives), a corrector would save a lot of time that would otherwise have been spent on finding those "needles in a haystack." Whereas *objective recall* is bounded by the number of OCR tokens with more than three characters, limiting it effectively to about 67% (there is quite a number of short words to start with, we count hyphenated words at line end as two different tokens, and a lot of split tokens contribute to this class as well: 401 tokens are split in 1557-W and 21 in 1609-K), one might ask what kind of tokens prevent *fair recall* to reach 100%. This difference are just the tokens labeled as *false negatives*. With method 0, this list contains a lot of uninterpretables; once these have been removed to positives, the list is dominated by *false friends*: exact matches against lexical entries which happen to be the wrong word at this position. Examples are the OCR tokens *hohe* (ground truth: *hóhe*), *find (find)*, or *zerteile (erteile)*. There are also *historical false friends*, matches against real or hypothesized historical wordforms not representing the ground truth such as *irer (jrer)*, *erfart (erfarēt)*, or *todt (tódt)*. A straightforward method for further improving fair recall (inspecting a larger number of tokens) would be to look not only at the *best-ranked* profiler interpretations, but to check all tokens with some interpretation as an error.

As word splits and merges are among the most frequent OCR errors, a large number of false negative tokens arise from word splits (some also from merges) that happen to be separately interpretable: *urfarbe (purpurfarbe)*, *búch (búchlein)*, or an example for a merge: *ober-farbe (ober farbe)*.

Whereas nothing can be done for original short tokens and not much about false friends, it would help a lot if merges and splits arising from the OCR process could be detected and remedied as part of the postcorrection process. If two consecutive tokens are not lexical, one could try if their concatenation is lexical. Perhaps external knowledge such as confidence values output by the OCR engine can also be used to detect possible splits whenever an interword space has a low confidence. Likewise, merges may be detectable by looking at the horizontal character coordinate indicating that there must have been an additional character in between originally (a whitespace) that got lost in recognition.

## 7. CONCLUSION

In this paper we introduced three methods for improving profiler-based error detection in OCR'ed historical texts. Evaluation results where given that show the positive effect of all methods suggested. The maximal reachable level of word accuracy by a purely profiler-based postcorrection can be calculated assuming that all profiler-detected error candidates get corrected. Word accuracy would then rise from 73% to 87% (1557-W) and from 91% to 96% (1609-K). Considering only tokens with length greater than 3 characters, the maximum *fair word accuracy* is at 96.6% and 98.5%, respectively, as only a handful of false friends go undetected. To reach these levels the baseline profiler methods are sufficient, but adaptivity reduces the amount of tokens to inspect to 27% and 17% of the total number of tokens (purely manual correction). As many of these tokens appear more than once or contain similar error patterns, corrections via PoCoTo would provide an additional efficiency boost.

The biggest obstacle in getting even better recall values are token splits and merges. We outlined lexical and external methods to detect them.

**Acknowledgements.** This work was partially funded by Deutsche Forschungsgemeinschaft (DFG) under grant no. SCHU-1026/7-1 (FF, US) and LU 856/7-1 (US).